%% file: pace_arxiv.tex
\documentclass{article} 
\usepackage{iclr2027_conference,times}

\input{math_commands.tex}

\usepackage{hyperref}
\usepackage{url}
\usepackage{booktabs}       
\usepackage{amsfonts}       
\usepackage{nicefrac}       
\usepackage{microtype}      
\usepackage{xcolor}         
\usepackage{graphicx}
\usepackage{multirow}
\usepackage{tcolorbox}
\usepackage{algorithm}
\usepackage{algorithmic}
\usepackage{amsmath}

\title{An Evolutionary Agentic Approach for Open-ended Image Quality Perception}

\newcommand{\codeurl}{https://github.com/2kxx/PACE}
\newcommand{\siteurl}{https://2kxx.github.io/PACE/}

\newcommand{\corremail}{}

\author{%
\begin{tabular}{c}
\begin{tabular}{cccc}
Zhenchen Tang\textsuperscript{1,2} & Bo Peng\textsuperscript{1,2,$\dagger$} &
Zichuan Wang\textsuperscript{1,2} & Songlin Yang\textsuperscript{3} \\[3pt]
Leilei Cao\textsuperscript{4} & Fengjie Zhu\textsuperscript{4} &
Jing Dong\textsuperscript{1,2,$\dagger$}\ &
\end{tabular}\\[8pt]
{\small\normalfont
\parbox{0.94\textwidth}{\centering
\textsuperscript{1}New Laboratory of Pattern Recognition, Institute of Automation, Chinese Academy of Sciences, Beijing, China\\
\textsuperscript{2}School of Artificial Intelligence, University of Chinese Academy of Sciences, Beijing, China\\
\textsuperscript{3}The Hong Kong University of Science and Technology, Hong Kong SAR, China\\
\textsuperscript{4}Transsion, Shenzhen, China\\[2pt]
$\dagger$~Corresponding author\corremail
}}\\[6pt]
{\small\normalfont Code: \url{\codeurl}}\\[2pt]
{\small\normalfont Project page: \url{\siteurl}}
\end{tabular}}

\iclrfinalcopy 

\begin{document}

\maketitle
\lhead{} 

\begin{abstract}

Generative models are rapidly expanding image quality assessment (IQA) beyond traditional fidelity factors to emerging dimensions such as physical plausibility and text-rendering correctness. However, existing IQA models rely on fixed definitions and heavy supervision, making them difficult to extend to open-ended perceptual dimensions. We identify holistic bias as an important limitation: when scoring an unseen dimension, models reuse generic quality priors, leading to scoring errors and rank inversion. To address this, we propose PACE (Perceptual Agentic Collaborative Evolution), a training-free multi-agent framework that formulates open-ended IQA as explicit protocol construction. Given a target dimension, PACE uses collaborative agents to construct an evaluation protocol composed of verifiable Visual Question Answering (VQA) probes, grounding evaluation in concrete visual evidence rather than holistic impressions. The resulting protocol is calibrated using only four human-annotated images per dimension, while a dual-track scoring mechanism aligns model perception with human scoring scales. Across traditional IQA, structural fidelity, context-aware aesthetics, and newly defined open-ended dimensions, PACE consistently improves its MLLM backbone, achieving competitive performance across diverse IQA settings, and reduces the Holistic Override Rate (HOR) from 44.4\% to 8.6\%.

\end{abstract}

\section{Introduction}

With the rapid growth of AI-generated content (AIGC), image quality assessment (IQA) has expanded beyond traditional fidelity factors such as blur, noise, and compression artifacts. Recent benchmarks further evaluate whether generated images respect world knowledge and physical commonsense \citep{niu2025wise}, render correct visual text \citep{wei2025tiif}, and follow the reasoning implied by a prompt \citep{zhao2026envisioning, tang2026endogenous}. Such dimensions often emerge faster than dedicated benchmarks or annotations can be built. Modern IQA therefore needs to handle open-ended dimensions—subjective and context-dependent attributes such as tactile satisfaction and human vitality. Existing methods based on fixed quality definitions and large-scale annotations are difficult to extend to these rapidly changing requirements \citep{tang2024clip, gong2025onereward}.

\begin{figure}[t]
  \vspace{-0.5cm}
  \centering
  \includegraphics[width=1.0\linewidth]{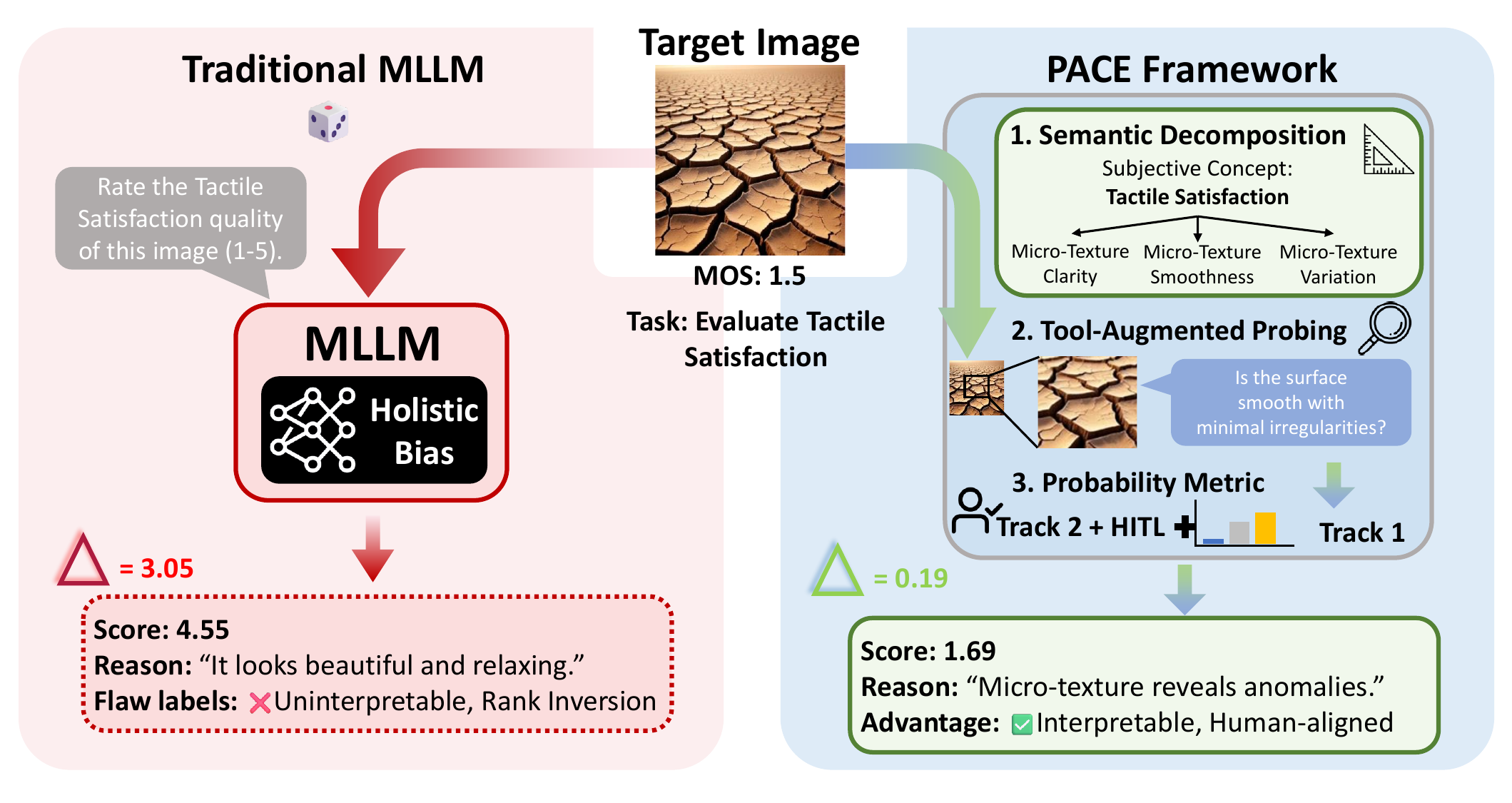}
  \vspace{-0.5cm}
  \caption{\textbf{Comparison between traditional MLLM scoring and PACE.} Direct MLLM scoring is affected by holistic bias, while PACE decomposes the target dimension (e.g., tactile satisfaction) into verifiable visual probes and produces evidence-based, human-aligned scores.}
  \label{fig:teaser}
  \vspace{-0.5cm}
\end{figure}

Recent multimodal large language models (MLLMs) provide a promising alternative because of their strong reasoning and semantic knowledge \citep{wu2024towards, wu2024q}. However, when directly applied to perceptual scoring, we observe a systematic failure mode that we term holistic bias. Specifically, when scoring an unseen dimension, MLLMs often reuse generic quality priors instead of evaluating the target dimension itself (see Fig. \ref{fig:teaser}). As a result, an image with high overall quality may receive a high dimension-specific score even when it fails badly on that dimension. We quantify this behavior using the Holistic Override Rate (HOR), which measures how often poor target-dimension images are incorrectly assigned high scores. 
As detailed in Sec.~\ref{sec:hor_quant} and Appendix~\ref{appendix:holistic_bias}, existing methods exhibit HOR values as high as 58.8\% on fine-grained dimensions such as tactile satisfaction. To reduce holistic bias, we decompose a subjective quality dimension into verifiable visual probes that together form an evaluation protocol. This forces the model to look at concrete visual details first, rather than making a quick guess based on general impressions. However, dynamically constructing these protocols for unseen dimensions is complex. To address this, we employ a collaborative multi-agent system to automate the process.

Specifically, we propose Perceptual Agentic Collaborative Evolution (PACE), a training-free multi-agent framework for open-ended IQA. Instead of directly predicting quality scores, PACE formulates perceptual evaluation as a protocol-construction process under a fast-and-slow thinking paradigm. For familiar dimensions, PACE reuses existing experts or previously evolved protocols. For unseen dimensions, a Planning Agent decomposes the target concept into observable sub-dimensions, while Visualizer and Critic agents refine them into verifiable VQA probes. A one-time human-in-the-loop (HITL) calibration then provides reusable scoring anchors. Finally, a dual-track scoring mechanism combines protocol-based scoring with human-aligned calibration. Beyond empirical performance, PACE provides a conceptual reframing of open-ended IQA: evaluation should be treated as on-the-fly protocol construction rather than static score prediction.

Our main contributions are summarized as follows:

\begin{itemize}
    \item \textbf{Autonomous Agentic System:} We propose PACE, to our knowledge the first multi-agent framework for open-ended IQA. PACE transforms subjective perceptual dimensions into measurable, reusable VQA-based evaluation protocols through collaborative agent evolution.
    \item \textbf{Efficient Human-Aligned Calibration:} We introduce a dual-track scoring algorithm paired with one-time human-in-the-loop (HITL) calibration, requiring only four annotated images per new dimension.
    \item \textbf{Broad Adaptability:} Across four evaluation tiers, PACE consistently improves its MLLM backbone, achieves competitive performance across diverse IQA settings, and reduces HOR from 44.4\% to 8.6\% on open-ended dimensions.
\end{itemize}

\section{Related Work}
\paragraph{Traditional Image Quality Assessment.}
Image quality assessment (IQA) has been extensively studied under both full-reference (FR) and no-reference (NR) settings. Traditional IQA methods model image quality using handcrafted features or learned models trained on large-scale MOS annotations \citep{hore2010image, wang2004image, mittal2012no, mittal2012making, bosse2017deep, talebi2018nima, su2020blindly, ke2021musiq, zhang2021uncertainty, zhang2023blind, wang2023exploring, tang2024clip}. While effective on standard benchmarks, these methods rely on predefined quality dimensions and dedicated annotations. Extending them to new perceptual dimensions therefore requires additional data and training, limiting their use in open-ended settings.

\paragraph{MLLM-based Perceptual Scoring.} Recent MLLMs provide flexible image scoring capabilities \citep{wu2023qb, wu2024q, wu2023q, you2025teaching, tang2026revisiting, zhou2025q}, but most directly predict final scores without explicit evaluation protocols. This makes them vulnerable to holistic bias on unseen dimensions. More importantly, their evaluation criteria remain implicit, making it difficult to verify whether a score is supported by target-specific visual evidence. Existing works explore open-ended comparison \citep{wu2024towards} or manually defined protocols \citep{tan2024evalalign}, but are limited by relative judgments or human effort. PACE instead automatically constructs reusable evaluation protocols for unseen dimensions with minimal human calibration.

\paragraph{Automated Evaluation and Multi-Agent Reasoning.}
Recent studies on LLM-as-a-judge have shown that complex judgments can benefit from explicit rubric and checklist-based evaluation \citep{gu2024survey,hashemi2024llm, gunjal2025rubrics, lee2025checkeval, zhang2026well, bai2026edit}. In parallel, multi-agent systems further improve reasoning through planning, critique, and collaboration \citep{wu2024autogen,du2024improving}. However, existing MLLM-based IQA methods still mainly assume predefined dimensions and direct score prediction. PACE instead uses collaborative agents to construct evaluation protocols for unseen dimensions.

\section{Holistic Bias: Definition and Quantification}
\label{sec:holistic_bias_def}

Before presenting PACE, we formalize the main failure mode that motivates our framework. We define \textbf{holistic bias} as the tendency of MLLMs to rely on overall image quality when scoring an unseen dimension, assigning high scores even when the target dimension is poor.

To quantify this behavior, we introduce the \textit{Holistic Override Rate (HOR)}. Let $N$ denote the number of evaluated images, $y_i \in [1, 5]$ the human MOS, and $\hat{y}_i \in [1, 5]$ the predicted score. Following the standard 5-point Likert scale, where scores below $3$ correspond to ``poor'' or ``bad'' quality and scores above $4$ indicate ``good'' or ``excellent'', we set the low-quality threshold $\tau_{low} = 3.0$ and the high-score threshold $\tau_{high} = 4.0$. The HOR is then defined as
\begin{equation}
    \mathrm{HOR} = \frac{\sum_{i=1}^{N} \mathbb{I}\left(y_i \le \tau_{low} \,\land\, \hat{y}_i \ge \tau_{high}\right)}{\sum_{i=1}^{N} \mathbb{I}(y_i \le \tau_{low})} \times 100\%,
    \label{eq:hor}
\end{equation}
where $\mathbb{I}(\cdot)$ is the indicator function. HOR measures the proportion of low-quality images that receive incorrectly high predictions. A higher HOR indicates stronger holistic bias. Unlike PLCC and SRCC, HOR focuses on a specific failure case where low target-quality images receive incorrectly high scores. Quantitative results are reported in Sec.~\ref{sec:hor_quant} and Appendix~\ref{appendix:holistic_bias}. To reduce holistic bias, PACE replaces direct scoring with explicit, evidence-based protocol construction, as detailed next.

\begin{figure}[t]
  \centering
  \includegraphics[width=\linewidth]{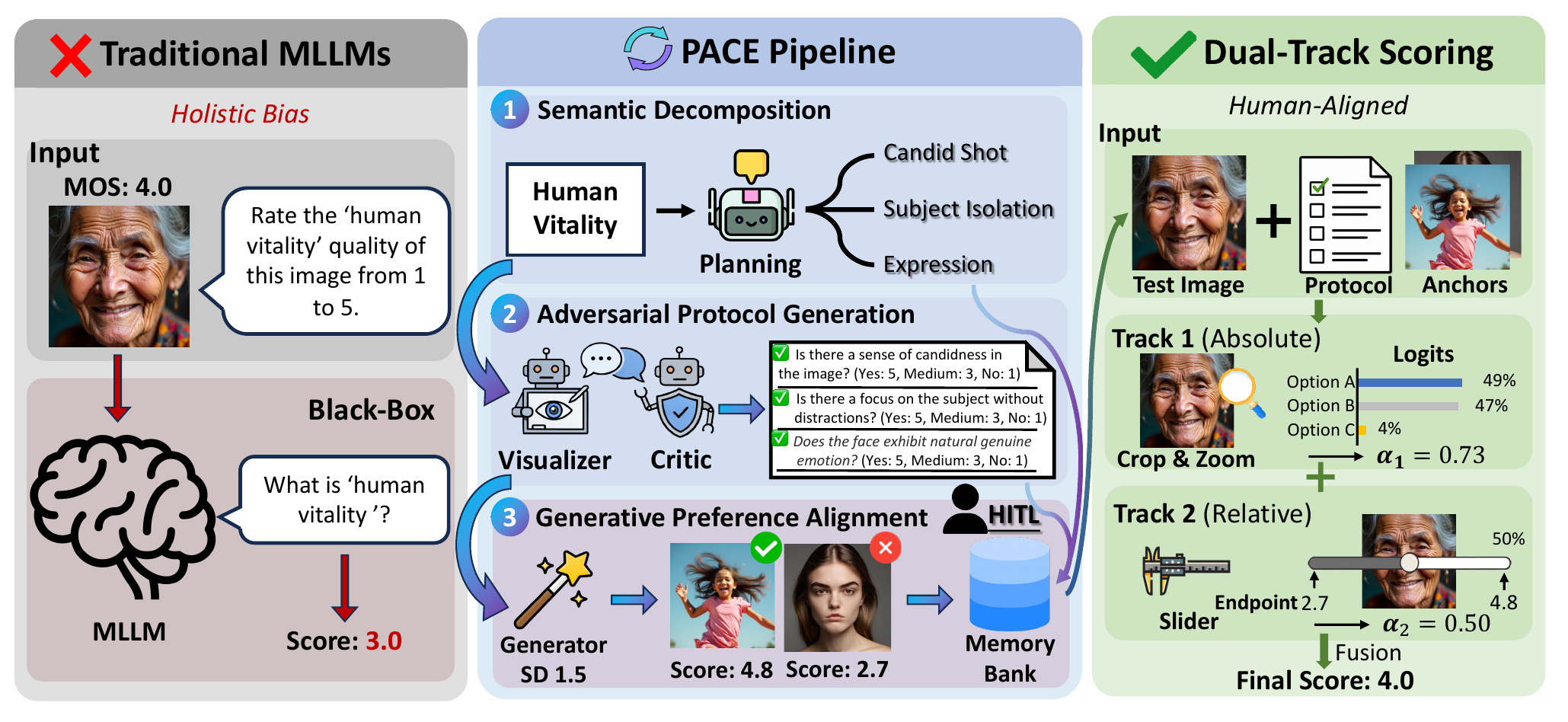}
  \vspace{-0.5cm}
  \caption{\textbf{Overview of the PACE pipeline and dual-track scoring.} (Left) The Planning Agent decomposes the target dimension into observable sub-dimensions, while the Visualizer and Critic construct and refine VQA probes. (Right) A Generator produces HITL calibration anchors. During scoring, Track 1 evaluates the image using the resulting evaluation protocol, while Track 2 aligns the prediction with the human-rated anchors. The two tracks are fused to produce the final score.}
  \label{fig:pipeline_scoring}
  \vspace{-0.5cm}
\end{figure}

\section{Methodology: The PACE Framework}
\label{methodology}
As illustrated in Fig. \ref{fig:pipeline_scoring}, PACE is a training-free multi-agent framework that formulates open-ended IQA as explicit protocol construction. Given an image and a target dimension, PACE constructs an evaluation protocol from visual evidence and maps the resulting perception to a human scoring scale. The framework contains four components: (1) fast–slow routing, (2) multi-agent protocol construction, (3) one-time HITL calibration, and (4) dual-track continuous scoring.

\subsection{Problem Formulation}
We study open-ended image quality assessment (IQA), where the goal is to predict a perceptual quality score for an image under an arbitrary, user-specified dimension. Formally, given a target image \(x\) and a textual quality dimension \(d\) (e.g., \textit{tactile satisfaction, human vitality}), the objective is to estimate a score \(y \in [1,5]\) that reflects human preference under dimension \(d\).

Unlike conventional IQA settings, the target dimension \(d\) may be previously unseen during system design, and no dedicated training annotations are assumed to be available. The key challenge is therefore to construct reliable evaluation criteria for new dimensions with minimal human supervision.

\subsection{Fast-and-Slow Thinking Paradigm}
To balance capability with computational cost, the Planning Agent employs a dynamic routing mechanism based on the familiarity of the target dimension $d$. Specifically, PACE routes each target dimension according to its similarity to previously evolved dimensions in the Retrieval-Augmented Generation (RAG) memory bank. We compute the cosine similarity between the embedding of $d$ and each known dimension $d'$. If $\max \text{cos}(\text{emb}(d), \text{emb}(d')) > \theta$, PACE uses the Fast-Thinking branch and reuses existing experts or evaluation protocols. Otherwise, it triggers the Slow-Thinking pipeline to construct a new protocol. Once calibrated, the new protocol is stored for future reuse. This avoids repeated protocol construction for familiar dimensions while preserving adaptability to unseen ones. We use all-MiniLM-L6-v2 \citep{wang2020minilm} to extract embeddings and set $\theta=0.8$. Further details are provided in Appendix~\ref{fastthinking}.

\subsection{Multi-Agent Perceptual Evolution}
We identify that direct MLLM scoring often suffers from holistic bias because generic quality priors are transferred to unseen dimensions. To mitigate this, PACE instead grounds evaluation in multiple target-specific visual probes. Constructing and refining such probes for unseen dimensions is complex, so we use specialized agents to collaboratively decompose the target concept, generate and refine visual probes, and construct calibration references: a Planning Agent, a Visualizer Agent, a Critic Agent, and a Generator Agent. Together, these roles separate what to evaluate, how to evaluate it, whether the resulting probes are valid, and how to construct human-aligned references.

\paragraph{Planning Agent: Concept Decomposition.}
Given a target dimension \(d\), the Planning Agent recursively decomposes it into a set of observable and minimally overlapping sub-dimensions:
\begin{equation}
\mathcal{S}(d)=\{s_1,s_2,\dots,s_n\}, \ n \in [3, 5].
\end{equation}
We use $n \in [3, 5]$ to balance coverage and inference cost: too few sub-dimensions may miss important aspects, while too many introduce redundant probes and additional inference. Each sub-dimension should be visually observable, specific, minimally overlapping, and jointly cover $d$. For example, human vitality may be decomposed into candidness, subject isolation, and facial expression.

\paragraph{Visualizer and Critic Agents: Protocol Construction.}
The Visualizer Agent transforms each sub-dimension into verifiable VQA probes with discrete scoring options. The provided image is used only as visual grounding to ensure that the generated probes are reasonable and visually relevant. The resulting probes capture dimension-level visual criteria rather than image-specific details, and can be reused across images under the same dimension. The granularity of the probes adapts to the target dimension: broad dimensions such as aesthetics use coarser criteria, while specific dimensions such as human vitality require finer visual details.

To ensure protocol reliability, the Critic Agent iteratively reviews the generated probes. Invalid probes are returned to the Visualizer for revision, with at most three rounds. The Critic rejects or corrects probes that are subjective or redundant, require information not visible in the image, or use an inverted score mapping where higher scores indicate worse quality. The accepted probe set forms an evaluation protocol for dimension \(d\), which is stored and reused for future evaluation.

\subsection{One-Time Human-in-the-Loop Calibration}
\label{HITL}
Although the evaluation protocol defines what should be measured, model outputs are not directly aligned with human scoring scales. PACE therefore performs a one-time human-in-the-loop (HITL) calibration for each new dimension.

A \textbf{Generator Agent }first converts the finalized evaluation protocol into positive and negative prompts, and then synthesizes four candidate images using a text-to-image model such as Stable Diffusion v1.5 \citep{rombach2022high}. Human annotators score all four images, and the highest-rated and lowest-rated samples are selected as two calibration anchors ($M=2$):
\begin{equation}
\mathcal{A}(d)=\{(a_m,y_m)\}_{m=1}^{M},
\end{equation}
where $a_m$ denotes an anchor image and $y_m \in [Y_{\text{low}}, Y_{\text{high}}]$ its human score. The anchors are stored in the memory bank and define the human-aligned scoring range for dimension $d$. The overall architecture of the PACE framework is provided in Fig. \ref{fig:framework} of Appendix~\ref{subsec:pseudo_code}.

\subsection{Dual-Track Continuous Scoring Algorithm}
To bridge the gap between machine perception and human evaluation, PACE introduces a novel dual-track scoring mechanism. Instead of directly predicting a final score, PACE uncouples the evaluation process into two distinct tracks: Track 1 focuses on \textit{Preference Alignment} by grounding the assessment in explicit visual evidence, while Track 2 ensures \textit{Score Alignment} by mapping the model perception to human scoring scales.

\paragraph{Track 1: Absolute Perception (Preference Alignment).} 
Track 1 evaluates the target image using the generated protocol. For a VQA probe $q_r$, let $C$ denote the candidate options (e.g., $\{Yes, Medium, No\}$) mapped to numerical quality scores $S_c$ (e.g., $\{5, 3, 1\}$, autonomously assigned by the Visualizer Agent during protocol generation). To perceive micro-features such as noise or film grain, PACE automatically invokes a tool-augmented probing mechanism (Crop \& Zoom) to provide local patches to the MLLM.
While extracting expected values from logits to create continuous score distributions has been explored in recent holistic IQA models (e.g., Q-Align), we adapt this technique to calculate the intensity of specific objective rules. We extract the raw logits $L_c$ (where $L_c$ denotes the logit corresponding to candidate option $c$)  from the MLLM's output and compute the Softmax probability distribution:
\begin{equation}
    P(c|I, q_r) = \frac{\exp(L_c)}{\sum_{c' \in C} \exp(L_{c'})}
\end{equation}
The expected perceptive score $E_r$ for rule $r$ is computed as the weighted sum:
\begin{equation}
    E_r = \sum_{c \in C} P(c|I, q_r) \cdot S_c \in [S_{min}, S_{max}]
\end{equation}
The absolute perceptive intensity $\alpha_{1} \in [0, 1]$ is subsequently obtained by aggregating all $N$ rules using uniform normalized weights $w_r = \frac{1}{N}$ in our implementation. To ensure the final intensity correctly maps to a standard probability-like interval, the weighted sum is normalized against $S_{max}$ and $S_{min}$, which represent the theoretical upper and lower bounds of the predefined candidate scores in $S_c$:
\begin{equation}
    \alpha_{1} = \frac{\sum_{r=1}^{N} w_r \cdot E_r - S_{min}}{S_{max} - S_{min}}
\end{equation}

\paragraph{Track 2: Relative Perception (Score Alignment).} 
Track 2 operates purely on comparative visuals to align the model perception with human scales. We compare the target image $I$ with the HITL-calibrated high-score anchor $A_{high}$ and low-score anchor $A_{low}$. The MLLM is then prompted to classify its relative position into 5 discrete ranks $K$, ranging from "Identical to $A_{high}$" to "Identical to $A_{low}$". The probability of rank $k$ is calculated from its corresponding logits $L_k$:
\begin{equation}
    P(k|I, A_{high}, A_{low}) = \frac{\exp(L_k)}{\sum_{k' \in K} \exp(L_{k'})}
\end{equation}
We assign $V_k \in \{1.0, 0.75, 0.5, 0.25, 0.0\}$ to the five ranks to smooth the discrete similarity levels and compute the continuous relative positioning coefficient $\alpha_{2}$:
\begin{equation}
    \alpha_{2} = \sum_{k \in K} P(k|I, A_{high}, A_{low}) \cdot V_k
\end{equation}

\paragraph{Final Score Fusion.} 
To yield the ultimate quality assessment, we fuse the absolute perception ($\alpha_{1}$) and the relative perception ($\alpha_{2}$) via balancing hyperparameters $\omega_1$ and $\omega_2$, where $\omega_1 + \omega_2 = 1$. We adopt a globally unified configuration ($\omega_1 = \omega_2 = 0.5$) for all experiments to avoid task-specific tuning, thereby ensuring PACE's robust applicability when generalizing to open-ended quality dimensions. The fused coefficient $\alpha_{final}$ is then linearly projected into the dynamic range defined by the calibration anchors $[Y_{low}, Y_{high}]$:
\begin{equation}
    \alpha_{final} = \omega_1 \cdot \alpha_{1} + \omega_2 \cdot \alpha_{2} 
\end{equation}
\begin{equation}
    Score_{final} = Y_{low} + (Y_{high} - Y_{low}) \cdot \alpha_{final}
\end{equation}
This dual-track linear projection mechanism expands the model’s score range and better aligns it with the human MOS distribution, enabling highly precise open-ended evaluation.

\section{Experiments}
\label{experiments}
We conduct a comprehensive evaluation to determine whether PACE can reliably assess both conventional and previously unseen perceptual dimensions. We organize our experiments into a \textbf{four-tier hierarchy}, ranging from traditional fidelity assessment to open-ended perception. All experiments are conducted in a training-free setting, reporting PLCC and SRCC as primary metrics.

\subsection{Experimental Setup}

PACE uses Qwen2.5-VL (7B) as the default multimodal backbone in the slow-thinking pipeline. The Planning, Visualizer, Critic, and Generator agents share the same backbone and differ in their system prompts and interaction context. To show that PACE is backbone-agnostic, we also evaluate it using mPLUG-Owl2 as an alternative backbone in Appendix~\ref{appendix:backbone_generalization}. For conventional dimensions, the fast-thinking mode additionally queries specialized IQA experts when available. We compare against three categories of baselines:

\begin{itemize}
    \item \textbf{Traditional IQA models:} NIQE \citep{mittal2012making}, BRISQUE \citep{mittal2012no}, NIMA \citep{talebi2018nima}, HyperIQA \citep{su2020blindly}, DBCNN \citep{zhang2018blind}, MUSIQ \citep{ke2021musiq}, CLIP-IQA+ \citep{wang2023exploring}.
    \item \textbf{Preference / Reward models:} PickScore \citep{kirstain2023pick}, ImageReward \citep{xu2023imagereward}, HPSv2 \citep{wu2023human}. These are selected for Tiers 2-4 as they are widely recognized for their strong zero-shot generalization in open-ended visual scoring.
    \item \textbf{MLLM-based IQA methods:}  Compare2Score \citep{zhu2024adaptive}, Q-Align \citep{wu2023q}, DeQA-Score \citep{you2025teaching}, Q-Scorer \citep{tang2026revisiting}, and a Qwen2.5-VL \citep{qwen25vl} baseline. To ensure a fair comparison, this Qwen2.5-VL baseline uses the same logit-weighting mechanism as Q-Align for holistic scoring. Furthermore, in Tiers 2–4, we design specialized, customized prompts for each target dimension for baseline models (e.g., Q-Align and Qwen2.5-VL), rather than using generic holistic prompts.
\end{itemize}

The four evaluation tiers are designed to progressively test: (1) standard fidelity robustness, (2) structure-sensitive reasoning, (3) context-aware aesthetics, and (4) generalization to unseen open-ended perceptual dimensions. In practical open-ended use, PACE follows the generative HITL calibration in Sec. \ref{HITL}, where human scores define the preference range of a new dimension. For benchmark evaluation, we select high-MOS and low-MOS images from the training split as anchors, directly using the benchmark annotations instead of collecting additional human scores. Calibration and test splits are strictly disjoint, and the anchors only define the scoring range without updating model parameters or protocols.

\subsection{Tier 1: Traditional IQA Benchmarks}

We first evaluate PACE on standard synthetic and in-the-wild IQA benchmarks, including KonIQ-10k \citep{hosu2020koniq}, SPAQ \citep{fang2020perceptual}, KADID-10k \citep{lin2019kadid}, LIVE-Wild \citep{ghadiyaram2015live}, AGIQA-3K \citep{li2023agiqa}, and CSIQ \citep{larson2010most}. This tier tests whether a framework designed for open-ended perception can remain competitive on conventional quality assessment tasks.
Note that baselines are trained on KonIQ. To ensure a fair comparison, we explicitly separate our results: Ours (Fast) utilizes existing expert models merely for routing efficiency on known domains, whereas the true capability of our protocol-construction pipeline is demonstrated by Ours (Slow). Operating entirely training-free, Ours (Slow) still significantly outperforms the direct Qwen2.5-VL baseline across all benchmarks.

Results are shown in Tab. \ref{tab:tradition}. PACE (Fast) preserves strong performance on established IQA benchmarks by reusing specialized experts and previously evolved protocols (e.g., 0.962 PLCC on KonIQ and 0.930 on SPAQ). Operating without any dataset-specific training, the autonomous multi-agent protocol-construction pipeline achieves remarkable performance (e.g., 0.782 PLCC on KADID). It significantly surpasses its direct Qwen2.5-VL backbone (0.576) by substantial margins. These results show the benefit of protocol-based evaluation over direct scoring.

\begin{table*}[t]
\centering
\caption{\textbf{Performance comparison on traditional and in-the-wild IQA benchmarks (Tier 1).} Results are evaluated by PLCC / SRCC. The best results are highlighted in \textbf{bold}.}
\label{tab:tradition}
\resizebox{\textwidth}{!}{
\begin{tabular}{clcccccc}
\toprule
\textbf{Category} & \textbf{Methods} & \textbf{KonIQ} & \textbf{SPAQ} & \textbf{KADID} & \textbf{LIVE-Wild} & \textbf{AGIQA-3K} & \textbf{CSIQ} \\
\midrule
\multirow{2}{*}{Handcrafted} 
& NIQE    & 0.533 / 0.530 & 0.679 / 0.664 & 0.468 / 0.405 & 0.493 / 0.449 & 0.560 / 0.533 & 0.718 / 0.628 \\
& BRISQUE & 0.225 / 0.226 & 0.490 / 0.406 & 0.429 / 0.356 & 0.361 / 0.313 & 0.541 / 0.497 & 0.740 / 0.556 \\
\midrule
\multirow{4}{*}{\shortstack{Non-MLLM\\Deep-learning}} 
& NIMA      & 0.896 / 0.859 & 0.838 / 0.856 & 0.532 / 0.535 & 0.814 / 0.771 & 0.715 / 0.654 & 0.695 / 0.649 \\
& HyperIQA  & 0.917 / 0.906 & 0.791 / 0.788 & 0.506 / 0.468 & 0.772 / 0.749 & 0.702 / 0.640 & 0.752 / 0.717 \\
& DBCNN     & 0.884 / 0.875 & 0.812 / 0.806 & 0.497 / 0.484 & 0.773 / 0.755 & 0.730 / 0.641 & 0.586 / 0.572 \\
& MUSIQ     & 0.924 / 0.929 & 0.868 / 0.863 & 0.575 / 0.556 & 0.789 / 0.830 & 0.722 / 0.630 & 0.771 / 0.710 \\
& CLIP-IQA+ & 0.909 / 0.895 & 0.866 / 0.864 & 0.653 / 0.654 & 0.832 / 0.805 & 0.736 / 0.685 & 0.772 / 0.719 \\
\midrule
\multirow{6}{*}{MLLM-based} 
& Compare2Score & 0.923 / 0.910 & 0.867 / 0.860 & 0.500 / 0.453 & 0.786 / 0.772 & 0.777 / 0.671 & 0.735 / 0.705 \\
& Q-Align       & 0.941 / 0.940 & 0.886 / 0.887 & 0.674 / 0.684 & 0.853 / 0.860 & 0.772 / 0.735 & 0.785 / 0.737 \\
& DeQA-Score    & 0.953 / 0.941 & 0.895 / 0.896 & 0.694 / 0.687 & 0.892 / 0.879 & 0.809 / 0.729 & 0.787 / 0.744 \\
& Q-Scorer      & 0.959 / 0.948 & 0.898 / 0.898 & 0.676 / 0.671 & 0.889 / 0.870 & 0.821 / 0.736 & 0.796 / 0.746 \\
& Qwen2.5-VL    & 0.737 / 0.692 & 0.855 / 0.860 & 0.576 / 0.522 & 0.625 / 0.615 & 0.813 / 0.744 & 0.724 / 0.679 \\
\cmidrule{2-8}
& \textbf{Ours (Slow)} & 0.841 / 0.783 & 0.900 / 0.893 & 0.782 / 0.776 & 0.807 / 0.760 & 0.821 / 0.758 & 0.735 / 0.697 \\
& \textbf{Ours (Fast)} & \textbf{0.962} / \textbf{0.950} & \textbf{0.930} / \textbf{0.929} & \textbf{0.926} / \textbf{0.922} & \textbf{0.907} / \textbf{0.892} & \textbf{0.822} / \textbf{0.751} & \textbf{0.880} / \textbf{0.835} \\
\bottomrule
\end{tabular}
}
\vspace{-0.5cm}
\end{table*}

\subsection{Tier 2: Structural and Logical Integrity}
We next evaluate whether PACE can detect local structural failures that are often missed by holistic scorers. We use HandEval \citep{wang2025handeval} (AI-generated hand anatomy realism) and PIPAL \citep{jinjin2020pipal}, which contains multiple restoration artifacts such as super-resolution failures, denoising errors, and GAN artifacts (see Appendix~\ref{appendix:dim_descriptions} for detailed descriptions of these dimensions).

Results in Tab. \ref{tab:structure} show that preference-oriented reward models perform poorly on these tasks, often approaching random correlation on HandEval. This failure illustrates holistic bias: visually attractive images may still contain severe local defects.
PACE substantially improves performance by explicitly decomposing structural quality into measurable protocols and inspecting local regions through crop-and-zoom probing. On HandEval, PACE achieves 0.575 SRCC, outperforming Qwen2.5-VL (0.501) and Q-Align (0.528). Similar gains are observed across PIPAL subcategories, especially GAN-based super-resolution and restoration tasks.
These results show that explicitly constructed protocols provide more reliable scoring on these structure-sensitive tasks.

\begin{table*}[t]
\centering
\caption{\textbf{Evaluation on Structural Integrity (Tier 2).} The results highlight the performance on complex biological and algorithmic hallucinations, including HandEval and PIPAL subtypes.}
\label{tab:structure}
\resizebox{\textwidth}{!}{
\begin{tabular}{lcccccc}
\toprule
\textbf{Methods} & \textbf{HandEval} & \textbf{PSNR-SR} & \textbf{GAN-based SR} & \textbf{Denoising} & \textbf{SR Full} & \textbf{PIPAL} \\
\midrule
CLIP-IQA     & 0.403 / 0.423 & 0.382 / 0.325 & 0.359 / 0.354 & 0.325 / 0.326 & 0.250 / 0.240 & 0.315 / 0.305 \\
PickScore    & 0.000 / 0.002 & 0.120 / -0.105 & 0.271 / -0.217 & 0.102 / 0.048 & 0.173 / -0.134 & 0.058 / -0.013 \\
ImageReward  & 0.282 / 0.279 & 0.244 / 0.129 & 0.232 / 0.189 & 0.231 / 0.182 & 0.160 / 0.132 & 0.163 / 0.130 \\
HPSv2        & 0.010 / 0.059 & 0.275 / 0.260 & 0.351 / 0.317 & 0.250 / 0.226 & 0.144 / 0.107 & 0.189 / 0.167 \\
Q-Align      & 0.521 / 0.528 & 0.438 / 0.430 & 0.515 / 0.494 & 0.426 / 0.389 & 0.407 / 0.408 & 0.403 / 0.419 \\
Qwen2.5-VL   & 0.503 / 0.501 & 0.471 / 0.442 & 0.425 / 0.390 & 0.346 / 0.304 & 0.416 / 0.351 & 0.390 / 0.357 \\
\midrule
\textbf{Ours (Slow)} & \textbf{0.572} / \textbf{0.575} & \textbf{0.607} / \textbf{0.592} & \textbf{0.536} / \textbf{0.520} & \textbf{0.442} / \textbf{0.407} & \textbf{0.496} / \textbf{0.448} & \textbf{0.493} / \textbf{0.457} \\
\midrule
\end{tabular}
}
\vspace{-0.3cm}
\end{table*}

\subsection{Tier 3: Context-Aware Aesthetic Adaptability}
Aesthetic quality is highly dependent on semantic context: criteria suitable for urban photography may not apply to flowers or ocean scenes. We therefore evaluate on five representative subsets of TAD66k \citep{he2022rethinking}: City, Nature, Sea, Yellow, and Flower (detailed in Appendix~\ref{appendix:dim_descriptions}).

As shown in Tab. \ref{tab:context}, PACE generally improves over static aesthetic scorers and direct MLLM baselines across the evaluated subsets. The gains are particularly clear in City and Yellow scenes.
We attribute this improvement to adaptive protocol construction. Instead of using a single holistic notion of aesthetics, PACE generates context-specific criteria for each domain. For example, geometric perspective and skyline balance may dominate City scenes, while color harmony and depth-of-field cues become more important for Flower scenes.
These results suggest that open-ended aesthetic judgment benefits from dynamically constructed criteria rather than fixed preference priors.

\begin{table*}[t]
\centering
\caption{\textbf{Evaluation on Context-Aware Aesthetics (Tier 3).} Performance across 5 representative scenes from the TAD66k dataset, demonstrating PACE's adaptability to diverse semantic contexts.}
\label{tab:context}
\resizebox{\textwidth}{!}{
\begin{tabular}{lccccc}
\toprule
\textbf{Methods} & \textbf{City} & \textbf{Nature} & \textbf{Sea} & \textbf{Yellow} & \textbf{Flower} \\
\midrule
CLIP-IQA     & 0.210 / -0.120 & 0.152 / 0.008 & 0.217 / -0.153 & 0.054 / 0.007 & 0.120 / 0.009 \\
PickScore    & 0.251 / 0.163  & 0.233 / 0.224 & 0.151 / 0.141 & 0.237 / 0.233 & 0.125 / 0.068 \\
ImageReward  & 0.217 / 0.140  & 0.220 / 0.165 & 0.115 / 0.062  & 0.268 / 0.219 & 0.131 / 0.049 \\
HPSv2        & 0.259 / 0.242  & 0.152 / 0.128 & 0.195 / 0.141  & 0.213 / 0.197 & 0.031 / 0.026 \\
Q-Align      & 0.304 / 0.285  & 0.118 / 0.129 & 0.170 / 0.110  & 0.228 / 0.202 & 0.157 / 0.086 \\
Qwen2.5-VL   & 0.391 / 0.370  & 0.295 / 0.290 & 0.249 / 0.206  & 0.213 / 0.225 & 0.144 / 0.111 \\
\midrule
\textbf{Ours (Slow)} & \textbf{0.438} / \textbf{0.409} & \textbf{0.327} / \textbf{0.279} & \textbf{0.278} / \textbf{0.239} & \textbf{0.293} / \textbf{0.263} & \textbf{0.169} / \textbf{0.157} \\
\bottomrule
\end{tabular}
}
\vspace{-0.3cm}
\end{table*}

\subsection{Tier 4: Zero-Training-Data Open-Ended Dimensions}
The most challenging setting in this work involves perceptual dimensions for which no dedicated training data exists. We define six representative open-ended dimensions (details of data construction and annotation are provided in Appendix~\ref{sec:appendix_open_ended_dataset}):

\begin{enumerate}
    \item \textbf{Text Rendering Fidelity (Typo):} Clarity and structural correctness of generated text.
    \item \textbf{Lighting Consistency (Light):} Physical coherence of ray tracing, shadows, and reflections.
    \item \textbf{Tactile Satisfaction (Tactile):} Smoothness and texture appealing to the sense of touch.
    \item \textbf{Surrealist Coherence (Dream):} Logical blending of materials in absurd or dream-like art.
    \item \textbf{Human Vitality (Portrait):} Authentic emotional expression, avoiding the uncanny valley.
    \item \textbf{Cinematic Narrative (Cine):} Dramatic framing and lighting that evoke a storytelling atmosphere.
\end{enumerate}

These dimensions are subjective, long-tail, and poorly covered by conventional IQA datasets. Results in Table~\ref{tab:open} reveal a substantial performance gap between direct scoring methods and PACE. General preference models frequently collapse or exhibit polarity inversion (negative SRCC and PLCC), indicating that generic quality priors override task-specific reasoning (holistic bias). For instance, on the Light dimension, ImageReward and HPSv2 exhibit severe negative PLCCs (-0.462 and -0.257). This provides clear evidence of holistic bias: the baseline models assign high scores to visually polished images even if the lighting logic is entirely flawed, inverting the true human preference. Furthermore, even the strong MLLM struggles on Typo and Light. PACE substantially reduces this inversion through explicit evaluation protocols.
PACE achieves the best overall performance with 0.714 PLCC and 0.666 SRCC, outperforming Qwen2.5-VL (0.493 / 0.501) and Q-Align (0.170 / 0.264). Strong gains are observed on Tactile Satisfaction and Human Vitality.

\paragraph{Quantitative Evidence of Holistic Bias.}
\label{sec:hor_quant} 
As shown in Fig. \ref{fig:hor_overall} and Fig. \ref{fig:hor_prominent}, the direct MLLM baseline suffers a catastrophic 44.4\% average HOR across unseen dimensions, spiking to 58.8\% on fine-grained tasks like Tactile Satisfaction. This is because holistic IQA models are easily distracted by holistic quality. When an image fails in a specific dimension but has high overall aesthetics, the model often inflates the dimensional score by relying on global priors rather than targeted evaluation. By judging images based on concrete evaluation protocols rather than a general impression, PACE substantially reduces this problem. It reduces the overall HOR to 8.6\%, and further achieves 2.9\% HOR on the challenging Tactile Satisfaction and Human Vitality subset. Tab. \ref{tab:hb_quality_evidence} in Appendix~\ref{appendix:holistic_bias} further shows that holistic models over-score these GT-low images (overall mean $\sim$3.1--3.6 vs. human GT $2.09$), further supporting the effect of holistic bias.

\begin{table*}[t]
\centering
\caption{\textbf{Zero-Training-Data Discovery of Open-Ended Dimensions (Tier 4).} Evaluation of 6 novel and subjective perceptual dimensions. Most baseline models experience polarity inversion (negative SRCC) due to holistic bias, whereas PACE maintains highly aligned continuous scoring.}
\label{tab:open}
\resizebox{\textwidth}{!}{
\begin{tabular}{lccccccc}
\toprule
\textbf{Methods} & \textbf{Typo} & \textbf{Light} & \textbf{Tactile} & \textbf{Dream} & \textbf{Portrait} & \textbf{Cine} & \textbf{OVERALL} \\
\midrule
CLIP-IQA     & -0.142 / -0.240 & 0.128 / 0.051 & 0.095 / 0.090 & 0.430 / 0.363 & 0.423 / 0.416 & 0.624 / 0.637 & 0.260 / 0.219 \\
PickScore    & -0.073 / -0.028 & 0.043 / 0.015 & 0.008 / -0.002 & 0.675 / 0.595 & 0.225 / 0.159 & 0.611 / 0.649 & 0.248 / 0.231 \\
ImageReward  & -0.104 / -0.176 & -0.462 / -0.205 & 0.071 / 0.116 & 0.746 / 0.661 & -0.294 / -0.216 & 0.736 / 0.714 & 0.115 / 0.149 \\
HPSv2        & -0.013 / 0.062 & -0.257 / -0.161 & 0.388 / 0.437 & 0.697 / 0.725 & -0.010 / -0.038 & 0.617 / 0.609 & 0.237 / 0.272 \\
Q-Align      & 0.461 / 0.608 & -0.187 / -0.211 & 0.312 / 0.386 & 0.009 / 0.155 & -0.058 / 0.125 & 0.487 / 0.519 & 0.170 / 0.264 \\
Qwen2.5-VL   & 0.339 / 0.480 & -0.088 / -0.057 & 0.604 / 0.644 & \textbf{0.874} / 0.758 & 0.535 / 0.468 & 0.694 / 0.710 & 0.493 / 0.501 \\
\midrule
\textbf{Ours (Slow)} & \textbf{0.684} / \textbf{0.673} & \textbf{0.348} / \textbf{0.204} & \textbf{0.838} / \textbf{0.796} & 0.849 / \textbf{0.785} & \textbf{0.806} / \textbf{0.790} & \textbf{0.759} / \textbf{0.745} & \textbf{0.714} / \textbf{0.666} \\
\bottomrule
\end{tabular}
}
\vspace{-0.5cm}
\end{table*}

\subsection{Ablation Studies}
As shown in Tab. \ref{tab:ablation}, we verify the effectiveness of PACE's main components. The Qwen2.5-VL baseline directly predicts quality scores, while Track 2 Only removes the protocol-construction branch and Track 1 Only removes anchor-based relative scoring. Both tracks improve over the baseline and provide complementary benefits. Specifically, Track 1 demonstrates superior performance on synthetic benchmarks like KADID (0.779), highlighting its strength in fine-grained evidence perception, while Track 2 effectively aligns score scales with human perception. Their balanced fusion ($\omega_1=0.5, \omega_2=0.5$) achieves the most stable performance across benchmarks, notably reaching 0.900 PLCC on SPAQ. 

For the multi-agent design, removing the protocol-construction branch clearly reduces performance, and removing the Critic causes a further drop. These results show that role-based collaboration and iterative refinement are important for reliable protocol construction. Since all agents share the same Qwen2.5-VL backbone, these gains come from role separation and interaction rather than additional model capacity.

\begin{table*}[t]
\centering
\caption{\textbf{Ablation studies of PACE components.} We analyze the effects of the dual-track scoring mechanism and multi-agent protocol construction.}
\label{tab:ablation}
\resizebox{\textwidth}{!}{
\begin{tabular}{lcccccc}
\toprule
\textbf{Methods / Config} & \textbf{KonIQ} & \textbf{SPAQ} & \textbf{KADID} & \textbf{LIVE-Wild} & \textbf{AGIQA-3K} & \textbf{CSIQ} \\
\midrule
Qwen2.5-VL (Q-Align)   & 0.737 / 0.692 & 0.855 / 0.860 & 0.576 / 0.522 & 0.625 / 0.615 & 0.813 / 0.744 & 0.724 / 0.679 \\
Track 2 Only  & 0.826 / 0.792 & 0.881 / 0.886 & 0.646 / 0.607 & 0.793 / 0.760 & 0.814 / 0.759 & 0.680 / 0.617 \\
Without Critic  & 0.836 / 0.792 & 0.888 / 0.880 & 0.761 / 0.756 & 0.792 / 0.751 & 0.807 / 0.742 & 0.705 / 0.663 \\
\midrule
Track 1 Only  & 0.798 / 0.738 & 0.868 / 0.860 & 0.779 / 0.775 & 0.763 / 0.709 & 0.778 / 0.710 & 0.729 / 0.695 \\
Ours ($\omega_1=0.3, \omega_2=0.7$) & \textbf{0.843} / \textbf{0.795} & 0.895 / 0.889 & 0.768 / 0.761 & \textbf{0.811} / \textbf{0.770} & \textbf{0.824} / \textbf{0.765} & 0.723 / 0.680 \\
Ours ($\omega_1=0.7, \omega_2=0.3$) & 0.829 / 0.768 & 0.886 / 0.877 & \textbf{0.783} / \textbf{0.780} & 0.793 / 0.744 & 0.807 / 0.739 & \textbf{0.737} / \textbf{0.703} \\
\textbf{Ours (slow, $\omega_1=0.5, \omega_2=0.5$)} & 0.841 / 0.783 & \textbf{0.900} / \textbf{0.893} & 0.782 / 0.776 & 0.807 / 0.763 & 0.821 / 0.758 & 0.735 / 0.697 \\
\bottomrule
\end{tabular}
}
\vspace{-0.5cm}
\end{table*}

\section{Conclusion}

In this paper, we introduced PACE, a training-free multi-agent framework for open-ended image quality assessment. By shifting the paradigm from direct score prediction to explicit protocol construction, PACE effectively reduces holistic bias. By combining semantic decomposition, adversarial protocol refinement, and one-time generative HITL calibration, PACE successfully navigates the diverse long-tail perceptual demands of the open world. Furthermore, our proposed dual-track scoring mechanism elegantly bridges the gap between objective machine probabilities and subjective human evaluation scales. Extensive experiments validate that PACE not only remains competitive on traditional benchmarks but also achieves strong adaptability on previously unseen dimensions.

\bibliographystyle{iclr2027_conference}
\bibliography{cite}

\newpage
\appendix
\section{Appendix}

\subsection{Implementation Details}
\label{sec:implementation_details}

In this section, we provide the concrete implementation details of our framework to ensure full reproducibility, including the algorithmic pseudo-code and the exact prompt templates used for the multi-agent system. \textbf{Code:} \url{\codeurl} \qquad \textbf{Project page:} \url{\siteurl}

\subsubsection{Pseudo-Code of PACE}
\label{subsec:pseudo_code}

Algorithm \ref{alg:pace} summarizes the main programmatic flow of PACE after HITL calibration.
It clarifies the exact variable flow across the Fast-Slow Routing, Multi-Agent Evolution, and Dual-Track Scoring phases discussed in the main text. We set the fusion weights $\omega_1 = \omega_2 =0.5$ in all experiments.

\begin{algorithm}[h]
\caption{PACE: Open-ended IQA via Multi-Agent Evolution}
\label{alg:pace}
\begin{algorithmic}[1]
\REQUIRE Target dimension $d$, image $x$, calibration anchors $(x_{low}, x_{high})$
\ENSURE Quality score $y \in [1, 5]$

\STATE \textbf{// Phase 1: Fast-Slow Routing}
\IF{$d \in \mathcal{D}_{known}$}
    \STATE Retrieve evolved protocol $\mathcal{P}(d)$ from RAG
\ELSE
    \STATE \textbf{// Phase 2: Multi-Agent Evolution}
    \STATE $\mathcal{S}(d) \leftarrow \text{PlanningAgent}(d)$ \COMMENT{Decompose into sub-dims}
    \FOR{$s_i \in \mathcal{S}(d)$}
        \STATE $q_i \leftarrow \text{VisualizerAgent}(s_i, x)$ \COMMENT{Generate VQA probe}
        \FOR{$t = 1$ to $3$}
            \STATE $f_i \leftarrow \text{CriticAgent}(q_i)$ \COMMENT{Review the probe}
            \IF{$f_i$ indicates valid}
                \STATE \textbf{break}
            \ENDIF
            \STATE $q_i \leftarrow \text{VisualizerAgent}(s_i, x, f_i)$ \COMMENT{Refine with feedback}
        \ENDFOR
    \ENDFOR
    \STATE $\mathcal{P}(d) \leftarrow \{(s_i, q_i)\}_{i=1}^{n}$
\ENDIF

\STATE \textbf{// Phase 3: Dual-Track Scoring}
\STATE $\alpha_1 \leftarrow \text{Track1\_PreferenceAlign}(x, \mathcal{P}(d))$
\STATE $\alpha_2 \leftarrow \text{Track2\_ScoreAlign}(x, x_{low}, x_{high})$
\STATE $\alpha_{final} \leftarrow \omega_1 \cdot \alpha_{1} + \omega_2 \cdot \alpha_{2}$
\STATE $y \leftarrow Y_{low} + \alpha_{final} \cdot (Y_{high} - Y_{low})$

\RETURN $y$
\end{algorithmic}
\end{algorithm}

\subsubsection{Prompt Templates for the PACE Agentic Framework}
\label{subsec:prompts}

To drive the Multi-Agent Evolution (\textbf{Phase 2} in Algorithm \ref{alg:pace}), we design strict system prompts to enforce role-playing and specific output formatting. The agents iteratively interact using these templates, where variables in brackets (e.g., \texttt{[dimension]}) are dynamically populated during runtime.

\begin{tcolorbox}[colback=blue!3!white,colframe=blue!70!black,title=Prompt 1: Planning Agent (Concept Decomposition), fonttitle=\bfseries]
I need to evaluate the visual quality dimension: '\texttt{[dimension]}'.

Please decompose this abstract concept into 3 to 5 distinct, visually observable sub-dimensions or low-level features.

Output JSON format: 

\texttt{\{ "sub\_dimensions": ["dim1", "dim2", ...] \}}
\end{tcolorbox}

\begin{tcolorbox}[colback=red!3!white,colframe=red!70!black,title=Prompt 2: Critic Agent (Visual Grounding Check), fonttitle=\bfseries]
You are a meta-evaluator for Visual Question Answering (VQA).

Please review the following proposed evaluation rule:

Question: "\texttt{[question]}"

Scoring mapping: "\texttt{[option scores]}"

Task: Is this a valid, visually grounded question that a Vision-Language Model can reasonably answer by observing an image?

- VALID: evaluates physical features (sharpness, noise) or visual semantics (style, realism).

- INVALID: purely emotional or requires invisible context.

CRITICAL: Evaluate both the question and its scoring mapping. Do not reject just because the image is low quality.

Reply STRICTLY in this format:

Yes. (or No.)

Reason: [short reason]
\end{tcolorbox}

\begin{tcolorbox}[colback=green!3!white,colframe=green!60!black,title=Prompt 3: Visualizer Agent (Protocol Generation), fonttitle=\bfseries]
You are an expert in Visual Quality Assessment and VQA Logic Design.

Your task is to create CONTEXT-AWARE and REUSABLE VQA rules for evaluating '\texttt{[dimension]}' with the provided image as visual grounding.

Identify visual elements, artifacts, or strengths related to \texttt{[sub\_dim\_str]}.
Your questions should focus on concrete details visible in the image.

\texttt{[ref\_context]}

Critical feedback from previous round:
"\texttt{[feedback or "None. This is the first iteration."]}"

REQUIREMENTS:

1. Generate \texttt{[max\_rules]} distinct VQA rules in JSON.

2. Each rule must have a specific Question and a Scoring Mapping (option\_scores).

3. Questions must be objective and focus on concrete visual evidence.

4. Do NOT ask abstract questions like "Is it good?". Ask "Is the edge sharp?" or "Are the colors natural?".

OUTPUT FORMAT (Strict JSON):
\begin{verbatim}
{
    "rules": [
        {
            "sub_dimension": "one of the sub features",
            "question": "The VQA question",
            "option_scores": { "High/Yes": 5, "Medium": 3, 
                                              "Low/No": 1 },
            "weight": 1.0,
            "rationale": "Why this question is included"
        }
    ]
}
\end{verbatim}
\end{tcolorbox}

\begin{figure}[h]
  \centering
  \includegraphics[width=\linewidth]{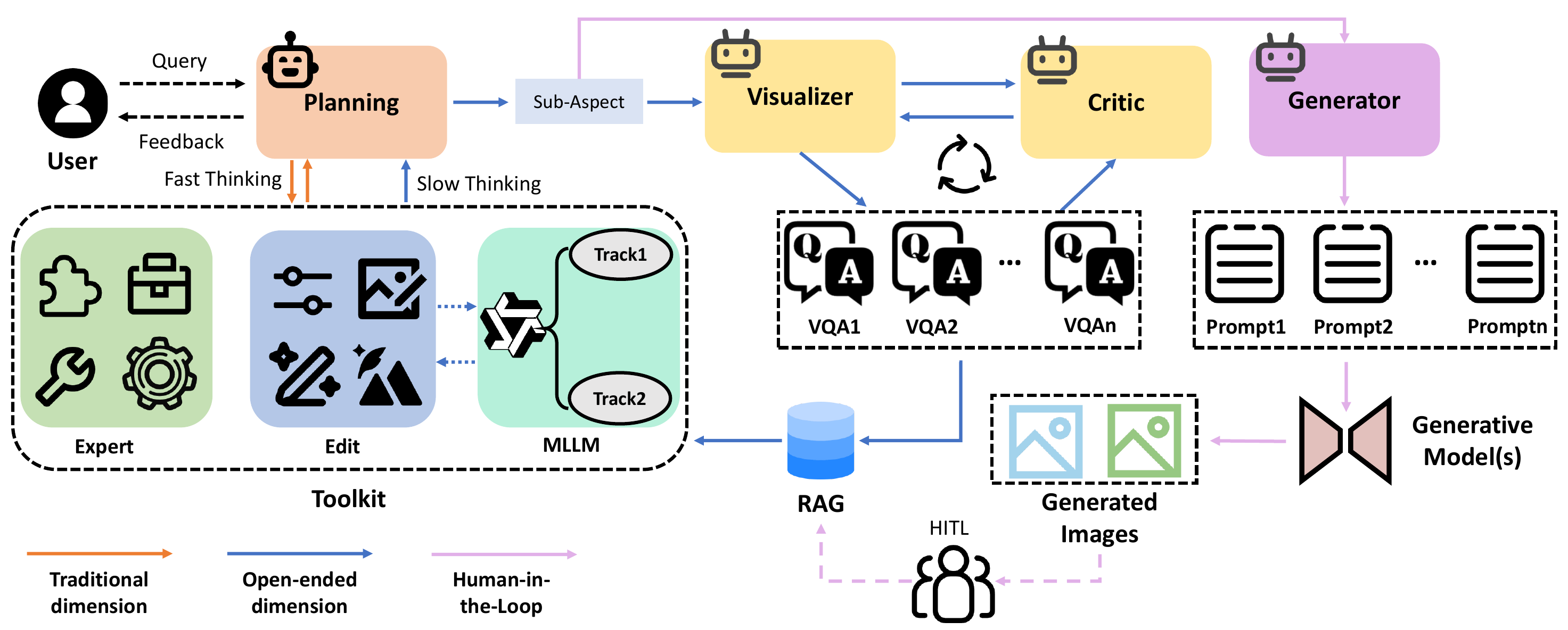} 
  \caption{\textbf{The overall architecture of the PACE framework.} Operating on a dual-process cognitive paradigm, PACE routes traditional dimensions to a fast thinking process via an expert toolkit. For open-ended dimensions, it triggers the slow thinking pipeline, where the Planning, Visualizer, Critic, and Generator agents collaborate to evolve VQA protocols and synthesize reference images. The knowledge is then solidified into a RAG memory bank for future rapid inference.}
  \label{fig:framework}
  \vspace{-0.5cm}
\end{figure}

\subsubsection{Fast-Thinking Branch Details.}
\label{fastthinking}
While the main text focuses on the Slow-Thinking pipeline, here we elaborate on the Fast-Thinking branch invoked when the cosine similarity between the target dimension and any known dimension exceeds the threshold $\theta=0.8$. When evaluating well-defined or previously evolved dimensions, the system bypasses the evolutionary process for high-throughput inference. For traditional fidelity dimensions (e.g., blur, noise), PACE queries an ensemble of specialized IQA experts (e.g., Q-Scorer, DeQA-Score) to compute a consensus score. This branch acts strictly as an engineering upper-bound for efficiency. For dimensions that have been previously evolved, PACE retrieves the solidified VQA protocols and their corresponding anchor image embeddings from a Vector RAG memory bank \citep{lewis2020retrieval}. By utilizing text-embedding cosine similarity to match new target dimension queries with previously evolved dimensions in the Vector RAG, the system bypasses the iterative Visualizer--Critic loop, significantly reducing inference latency for familiar dimensions.

\textit{Note: The Generator Agent constructs dynamic prompts for HITL calibration by feeding finalized VQA protocols into the MLLM to extract descriptive keywords, which are then formatted into standard positive/negative diffusion inputs.}

\subsubsection{Inference Runtime and Protocol Reuse}
\label{app:runtime}

We report wall-clock inference latency on the six Tier-4 open-ended dimensions. All measurements are conducted on a single NVIDIA A100 40GB GPU. For PACE (Slow), we measure the first image of each dimension, where the system performs concept decomposition, Visualizer--Critic refinement, and dual-track scoring. For PACE (Fast), we measure the next image after the evolved protocol has been stored and reused. Qwen2.5-VL and Q-Align are measured in separate runs using their corresponding task-aware scoring pipelines. Latency is reported in seconds per image.

\paragraph{Results.}
PACE (Slow) takes 22.72 s/image on average because it includes the one-time protocol evolution for a new dimension. Once the protocol is stored, PACE (Fast) reduces the latency to 0.67 s/image, corresponding to a $33.8\times$ speedup. Direct Qwen2.5-VL and Q-Align remain faster at 0.12 and 0.11 s/image because they directly produce quality scores without multi-agent protocol construction or multi-probe scoring. These results show that most of PACE's additional cost occurs when a new dimension is first introduced, while protocol reuse makes later evaluation substantially cheaper.

\subsection{Construction of Open-Ended Dimensions (Tier 4)}
\label{sec:appendix_open_ended_dataset}

To rigorously evaluate PACE on open-ended dimensions (Tier 4), we constructed a dedicated probing dataset consisting of 180 images across 6 novel dimensions, with 30 images for each dimension. The construction and annotation process followed a stringent three-step pipeline:

\begin{table}[h]
\centering
\caption{\textbf{Inference latency on the six Tier-4 open-ended dimensions.} Values are wall-clock seconds per image; lower is better. PACE (Slow) includes protocol evolution, while PACE (Fast) reuses the stored protocol.}
\label{tab:runtime}
\begin{tabular}{lccccccc}
\toprule
\textbf{Method} & \textbf{Typo} & \textbf{Light} & \textbf{Tactile} & \textbf{Dream} & \textbf{Portrait} & \textbf{Cine} & \textbf{Overall} \\
\midrule
PACE (Slow)   & 18.82 & 25.86 & 23.89 & 23.85 & 21.56 & 22.32 & 22.72 \\
PACE (Fast)   & 0.72  & 0.83  & 0.61  & 0.70  & 0.59  & 0.59  & 0.67 \\
Qwen2.5-VL    & 0.12  & 0.12  & 0.12  & 0.12  & 0.12  & 0.12  & 0.12 \\
Q-Align       & 0.13  & 0.13  & 0.11  & 0.10  & 0.11  & 0.11  & 0.11 \\
\bottomrule
\end{tabular}
\vspace{-0.5cm}
\end{table}

\paragraph{1. Concept Interpretation and Prompt Generation:} 
We leveraged the strong semantic comprehension capabilities of Gemini to interpret abstract perceptual dimensions (e.g., \textit{Human Vitality}, \textit{Tactile Satisfaction}). Through an iterative reprompting mechanism, we tasked Gemini to generate highly descriptive positive and negative text prompts that isolate the targeted dimension while maintaining overall image context.

\paragraph{2. Image Synthesis and Filtering:} 
Using the Gemini-generated prompts, we employed Bagel to synthesize an initial candidate pool of 60 images per dimension. Human experts manually inspected the pool and selected 30 representative images per dimension, removing samples with severe generation failures or artifacts unrelated to the target dimension. The selection was performed without access to any model predictions. The retained images span the perceptual quality spectrum, from severe failures to highly aligned manifestations.

\paragraph{3. Expert Human Annotation:} 
To establish the ground-truth Mean Opinion Scores (MOS), we invited 6 professional human annotators. The annotators were instructed to score the images on a 5-point scale (1 to 5) following a strict hierarchical rubric:
\begin{itemize}
    \item \textbf{Primary Criterion (Dimension Fulfillment):} Annotators must first evaluate how well the image satisfies the specific open-ended dimension (e.g., does the text rendering have structural correctness? Is the surrealist blending logically coherent?). If this primary task-specific criterion is severely flawed, the image must receive a low score (e.g., $\le 2$).
    \item \textbf{Secondary Criterion (Holistic Aesthetics):} Only when the primary criterion is reasonably met, annotators are allowed to fine-tune the score based on holistic visual factors, such as global aesthetic appeal, color harmony, and lighting quality. 
\end{itemize}
The final ground-truth MOS for each image is calculated by averaging the discrete scores from all 6 annotators.

\subsubsection{Inter-Annotator Reliability}
\label{app:annotation_reliability}

To verify the reliability of the Tier-4 human annotations, we evaluate the agreement among the six annotators for each open-ended dimension. Since the final ground-truth MOS is obtained by averaging the six ratings, we report ICC(2,$k$) under a two-way random-effects model with absolute agreement. We additionally report Cronbach's $\alpha$ as a complementary measure of rating consistency, together with the MOS and its 95\% confidence interval.

\begin{table}[t]
\centering
\caption{\textbf{Inter-annotator reliability on the Tier-4 open-ended dimensions.}
ICC(2,$k$) measures the reliability of the averaged ratings from six annotators. The 95\% confidence interval is reported for MOS.}
\label{tab:annotator_reliability}
\small
\begin{tabular}{lccc}
\toprule
\textbf{Dimension} & \textbf{ICC(2,$k$)} & \textbf{Cronbach's $\alpha$} & \textbf{MOS [95\% CI]} \\
\midrule
Text Rendering Fidelity & 0.911 & 0.919 & 3.08 [2.71, 3.46] \\
Lighting Consistency    & 0.675 & 0.677 & 3.69 [3.42, 3.93] \\
Tactile Satisfaction    & 0.868 & 0.872 & 2.32 [1.96, 2.71] \\
Surrealist Coherence    & 0.781 & 0.826 & 3.01 [2.69, 3.31] \\
Human Vitality          & 0.919 & 0.918 & 3.03 [2.62, 3.44] \\
Cinematic Narrative     & 0.652 & 0.655 & 3.33 [3.08, 3.59] \\
\midrule
\textbf{Overall}        & \textbf{0.859} & \textbf{0.862} & \textbf{3.08 [2.92, 3.22]} \\
\bottomrule
\end{tabular}
\vspace{-0.3cm}
\end{table}

\paragraph{Results.}
The annotations show stable reliability across all six open-ended dimensions. ICC(2,$k$) ranges from 0.652 to 0.919, while Cronbach's $\alpha$ ranges from 0.655 to 0.919. Four dimensions obtain ICC(2,$k$) values above 0.78, and the remaining two dimensions remain above 0.65. Across all 180 images, the overall ICC(2,$k$) reaches 0.859 and Cronbach's $\alpha$ reaches 0.862. A complementary Kendall's $W$ test also shows statistically significant agreement for every dimension ($p<0.01$). These results support the use of the averaged six-annotator MOS as a reliable human reference for the Tier-4 evaluation.

\subsection{Quantitative Evidence of Holistic Bias}
\label{appendix:holistic_bias}

This appendix expands on the formal definition of \textit{holistic bias} and the Holistic Override Rate (HOR) introduced in Sec.~\ref{sec:holistic_bias_def} of the main text by reporting the full experimental evidence and a complementary analysis on traditional quality entanglement.

\subsubsection{Experimental Results and Analysis}

\begin{figure}[t]
    \centering
    \includegraphics[width=0.95\textwidth]{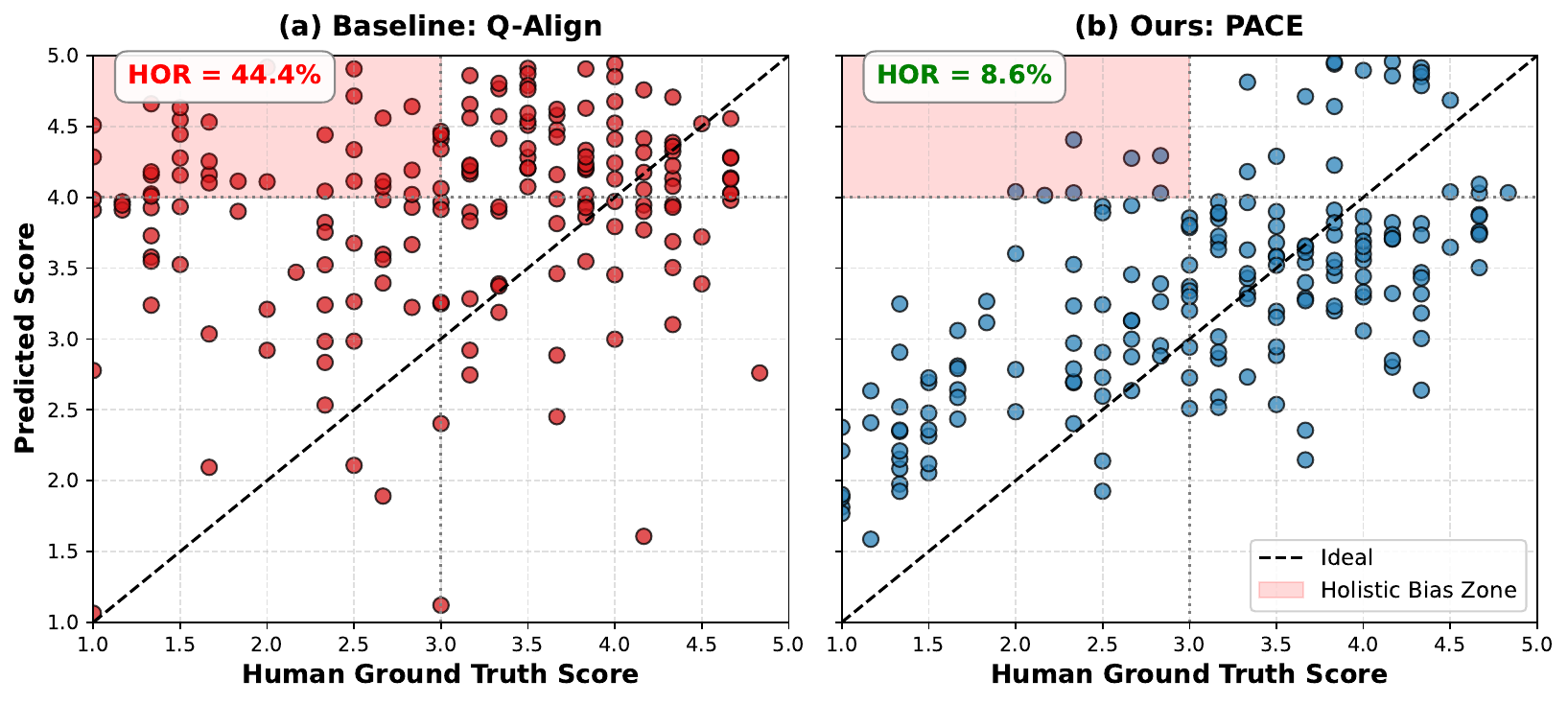}
    \caption{\textbf{Quantitative Evidence of Holistic Bias across All 6 Open-Ended Dimensions.} The red shaded area represents the \textit{Holistic Bias Zone} where human GT $\le 3.0$ but the model predicts $\ge 4.0$. (a) Q-Align frequently falls into this zone with a 44.4\% HOR. (b) PACE substantially reduces this bias, achieving an 8.6\% HOR.}
    \label{fig:hor_overall}
\end{figure}

\begin{figure}[t]
    \centering
    \includegraphics[width=0.95\textwidth]{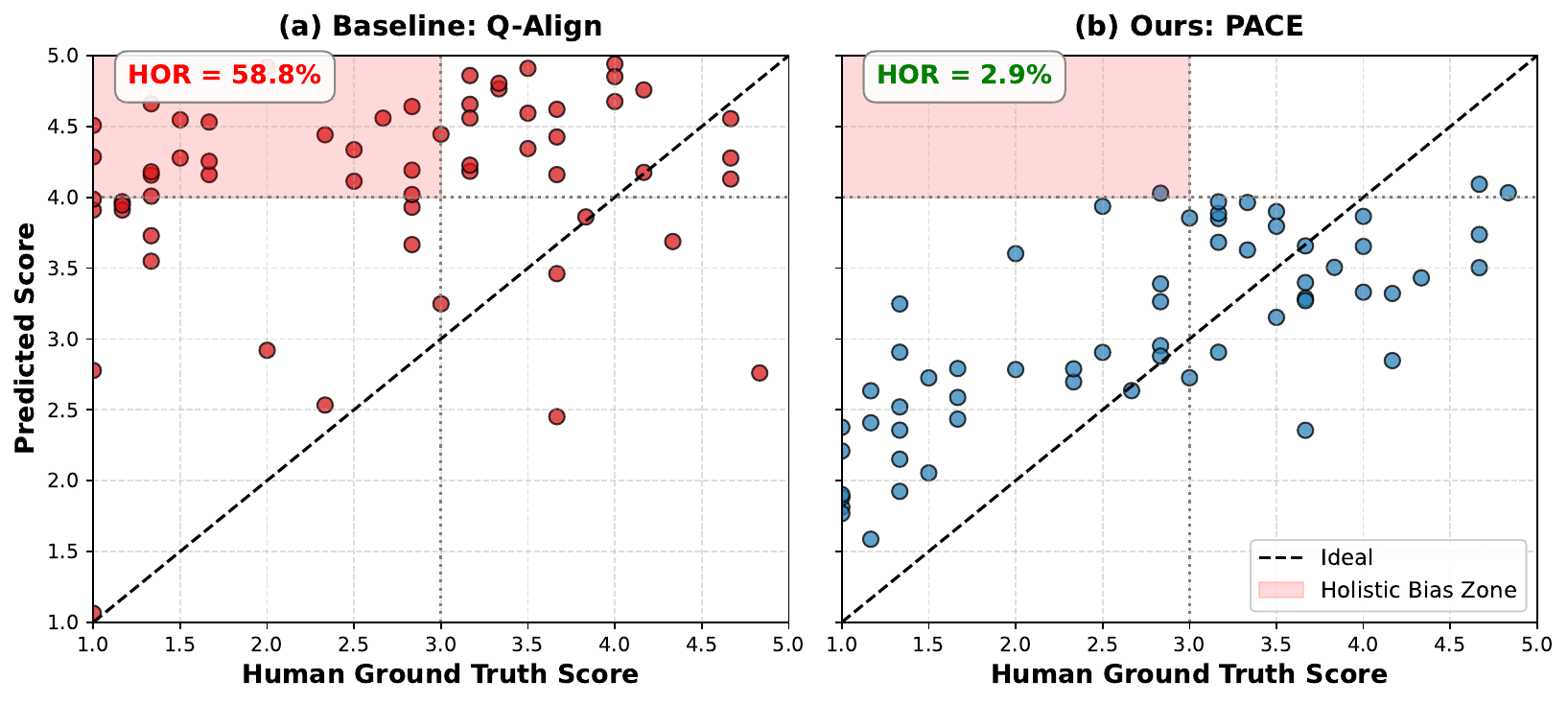}
    \caption{\textbf{Holistic Bias on Prominent Dimensions (\textit{Tactile Satisfaction} \& \textit{Human Vitality}).} On highly specific and challenging dimensions, direct MLLM prompting (Q-Align) suffers a catastrophic HOR of 58.8\%, showing strong interference from holistic aesthetic priors during task-specific reasoning. PACE maintains robust evidence-based scoring (2.9\% HOR).}
    \label{fig:hor_prominent}
\vspace{-0.5cm}
\end{figure}

We evaluated 180 images across 6 newly defined open-ended dimensions using Q-Align (a state-of-the-art MLLM-based holistic scorer) and our proposed PACE framework. The scatter plots are presented in Fig. \ref{fig:hor_overall} and Fig. \ref{fig:hor_prominent}.

\textbf{Failure on All Six Dimensions:} As illustrated in Fig. \ref{fig:hor_overall}, Q-Align demonstrates a severe holistic bias across all unseen dimensions. Out of 81 images with strictly poor human ratings ($y_i \le 3.0$), Q-Align erroneously assigned scores $\ge 4.0$ to 36 images, resulting in an overall HOR of \textbf{44.4\%}. In contrast, by grounding evaluation on granular, verifiable visual evidence, PACE successfully reduces this aesthetic interference, achieving an HOR of \textbf{8.6\%}.

\textbf{Failure on Prominent Dimensions:} The bias becomes catastrophically evident when evaluating open-ended dimensions such as \textit{Tactile Satisfaction} and \textit{Human Vitality}. As shown in Fig. \ref{fig:hor_prominent}, Q-Align entirely loses its discriminative ability on these specific dimensions. Out of 34 severely flawed images, Q-Align hallucinated high scores for 20 of them, raising the HOR to a staggering \textbf{58.8\%}. Conversely, PACE robustly anchors its predictions to human consensus, maintaining \textbf{2.9\%} HOR, showing the benefit of collaborative protocol construction over direct score prediction.
While dimensions like cinematic narrative naturally couple with aesthetics, our selected dimensions in Fig. \ref{fig:hor_prominent} (Tactile Satisfaction and Human Vitality) are decoupled from general aesthetics. Table~\ref{tab:hb_quality_evidence} confirms this: these images have low target quality but high holistic ratings. Despite the aesthetic distraction, PACE successfully identifies their defects using concrete probing protocols, avoiding the confusion that biases holistic models.

\subsubsection{Do Low-Quality Images Look Good in General?}
\label{subsec:hb_traditional_quality}

Why do holistic models fail on these specific dimensions? We find that images with low scores in a specific dimension actually look very clean and pretty in general. To test this, for each of the 81 GT-low images on Tier 4 ($y_i \le 3.0$ on the target open-ended dimension $d$), we additionally collect their predictions from three representative holistic scorers: \textbf{Q-Align}~\citep{wu2023q} (the baseline in our HOR analysis), \textbf{Q-Scorer}~\citep{tang2026revisiting} (a state-of-the-art MLLM-based IQA model), and \textbf{DeQA-Score}~\citep{you2025teaching} (a mPLUG-Owl2-based holistic quality scorer). These models are all trained for holistic image quality and do not know about the specific dimension $d$. We include Q-Align here to demonstrate that models producing high error rates on specific dimensions (Sec.~\ref{sec:hor_quant}) also assign high holistic scores to these same images, confirming the strong correlation between holistic aesthetic interference and targeted failures.

As shown in Tab. \ref{tab:hb_quality_evidence}, all holistic scorers systematically overrate these images: while human GT and PACE keep them on the low end of the scale (overall mean $2.09$ and $2.73$ respectively), Q-Align, Q-Scorer, and DeQA-Score push the same images well above their true quality, with DeQA-Score in particular hovering near $4.0$ on Dream and Portrait. This score inflation shows that holistic models are vulnerable to holistic quality interference. When the overall quality is high, specific dimensional ratings are often dragged upward. PACE solves this by focusing strictly on specific visual details through independent protocols.

\begin{table*}[t]
\centering
\caption{\textbf{Traditional quality of GT-low images on Tier 4 (holistic quality vs. specific quality).} For the 81 images that have poor human ratings on the target dimension $d$, we report the mean predictions of three holistic IQA models (Q-Align / Q-Scorer / DeQA-Score) alongside PACE's predictions and human GT. Values are on the standard $[1,5]$ scale.}
\label{tab:hb_quality_evidence}
\resizebox{\textwidth}{!}{
\begin{tabular}{lcccccc}
\toprule
\textbf{Dimension} & \textbf{\#GT-low} & \textbf{Q-Align $\uparrow$} & \textbf{Q-Scorer $\uparrow$} & \textbf{DeQA-Score $\uparrow$} & \textbf{PACE on $d$ $\downarrow$} & \textbf{GT on $d$ $\downarrow$} \\
\midrule
Typo (Text Rendering)         & 15 & 3.13 & 2.99 & 3.21 & 2.61 & 2.09 \\
Light (Lighting Consistency)  & 7  & 3.72 & 2.96 & 3.54 & 3.26 & 2.55 \\
Tactile (Tactile Satisfaction)& 19 & 4.01 & 3.07 & 3.67 & 2.66 & 1.61 \\
Dream (Surrealist Coherence)  & 12 & 4.20 & 3.30 & 3.97 & 2.69 & 2.11 \\
Portrait (Human Vitality)     & 15 & 4.25 & 3.32 & 3.98 & 2.68 & 2.09 \\
Cine (Cinematic Narrative)    & 13 & 3.36 & 2.68 & 3.14 & 2.80 & 2.54 \\
\midrule
\textbf{Overall (GT-low subset)} & \textbf{81} & 3.75 & \textbf{3.06} & \textbf{3.59} & \textbf{2.73} & \textbf{2.09} \\
\bottomrule
\end{tabular}
}
\vspace{-0.5cm}
\end{table*}

\subsection{Descriptions of Tier 2 and Tier 3 Evaluation Dimensions}
\label{appendix:dim_descriptions}

In this section, we provide detailed descriptions of the evaluation dimensions used in our Tier 2 and Tier 3 experiments. All dimensions are adopted from their original papers.

\subsubsection{Tier 2: Structural and Logical Integrity}
\label{appendix:tier2_dims}

We evaluate PACE on HandEval~\citep{wang2025handeval} and five evaluation configurations from PIPAL~\citep{jinjin2020pipal}:
\begin{itemize}
    \item \textbf{HandEval}: This dataset evaluates the perceptual quality of AI-generated human hands, focusing on structural correctness (e.g., finger count, proportions) and texture realism. We use it to test whether a model can identify local structural artifacts in generated content.
    \item \textbf{PIPAL} (full evaluation split): The complete evaluation set of PIPAL, which spans diverse distortion types including traditional distortions, super-resolution artifacts, and denoising results. It measures how well the model aligns with human ratings on a wide range of algorithmic distortions.
    \item \textbf{PSNR-SR}: This subset contains super-resolution results from models trained with PSNR/MSE loss. These images are often over-smoothed and lack fine details.
    \item \textbf{GAN-based SR}: This subset contains results from GAN-based super-resolution models. While they look sharp, they often hallucinate fake structures that do not match the ground truth.
    \item \textbf{Denoising}: This subset contains denoised images with residual noise or blur. It evaluates the model's sensitivity to the trade-off between noise removal and detail preservation.
    \item \textbf{SR Full}: This is the combination of all super-resolution subsets in PIPAL (i.e., PSNR-SR and GAN-based SR), evaluating the model's overall performance on super-resolution.
\end{itemize}

\subsubsection{Tier 3: Context-Aware Aesthetic Adaptability}
\label{appendix:tier3_dims}

The TAD66k dataset~\citep{he2022rethinking} contains 66,327 images covering 47 popular themes, where each theme is independently annotated with dedicated aesthetic criteria. To evaluate aesthetic quality fairly, we randomly select five themes covering diverse super-categories. This ensures a balanced and representative selection of contexts.

The five selected themes and their corresponding super-categories are:
\begin{itemize}
    \item \textbf{City}: Focuses on urban architecture, buildings, and city skylines.
    \item \textbf{Nature}: Focuses on natural scenery like mountains, forests, and fields.
    \item \textbf{Sea}: Focuses on ocean views, coastlines, and beach photography.
    \item \textbf{Yellow}: Focuses on images where yellow is the dominant color (e.g., autumn leaves or sunsets).
    \item \textbf{Flower}: Focuses on macro floral photography with soft background blur (depth-of-field).
\end{itemize}
These five themes cover diverse visual scenarios (large-scale scenes, color tones, and micro details), with each requiring distinct aesthetic standards.

\subsection{Generalization Across MLLM Backbones}
\label{appendix:backbone_generalization}

To demonstrate that PACE is backbone-agnostic, we evaluate its performance using mPLUG-Owl2~\citep{ye2024mplug} as an alternative backbone. We keep all other settings identical to the main experiments and compare against the direct holistic-scoring baseline on Tier 1 benchmarks.

\paragraph{Results.}
As shown in Table~\ref{tab:mplug_owl2_generalization}, PACE consistently improves the performance of mPLUG-Owl2 across five out of the six benchmarks. Specifically, on KonIQ, the PLCC/SRCC scores improve from $0.4290/0.3647$ to $0.5723/0.5265$. On LIVE-Wild, they increase from $0.5250/0.4430$ to $0.6326/0.5684$. On CSIQ, the scores increase from $0.5328/0.3585$ to $0.6317/0.5209$. We also observe stable gains on SPAQ and KADID.
The only exception is AGIQA-3K, where PACE shows a small drop in PLCC and unstable rank correlation. This is likely due to the known calibration instability of mPLUG-Owl2 on AIGC content. Overall, these results suggest that PACE can improve different MLLM backbones on most evaluated benchmarks.

\begin{table*}[t]
\centering
\caption{\textbf{Performance comparison of PACE using mPLUG-Owl2 backbone (Tier 1).} Values are PLCC / SRCC.}
\label{tab:mplug_owl2_generalization}
\resizebox{\textwidth}{!}{
\begin{tabular}{lcccccc}
\toprule
\textbf{Backbone / Method} & \textbf{KonIQ} & \textbf{SPAQ} & \textbf{KADID} & \textbf{LIVE-Wild} & \textbf{AGIQA-3K} & \textbf{CSIQ} \\
\midrule
mPLUG-Owl2 (baseline) & 0.4290 / 0.3647 & 0.7041 / 0.6775 & 0.4402 / 0.4343 & 0.5250 / 0.4430 & \textbf{0.5951} / \textbf{0.5826} & 0.5328 / 0.3585 \\
\textbf{mPLUG-Owl2 + PACE} & \textbf{0.5723} / \textbf{0.5265} & \textbf{0.7616} / \textbf{0.7466} & \textbf{0.4677} / \textbf{0.4344} & \textbf{0.6326} / \textbf{0.5684} & 0.5505 / 0.5453 & \textbf{0.6317} / \textbf{0.5209} \\
\midrule
$\Delta$ (Absolute) & +0.143 / +0.162 & +0.058 / +0.069 & +0.028 / +0.000 & +0.108 / +0.125 & -0.045 / -0.037 & +0.099 / +0.162 \\
\bottomrule
\end{tabular}
}
\end{table*}

\subsection{Qualitative Case Studies}
\label{appendix:qualitative_cases}

\begin{figure}[t]
    \centering
    \includegraphics[width=0.95\textwidth]{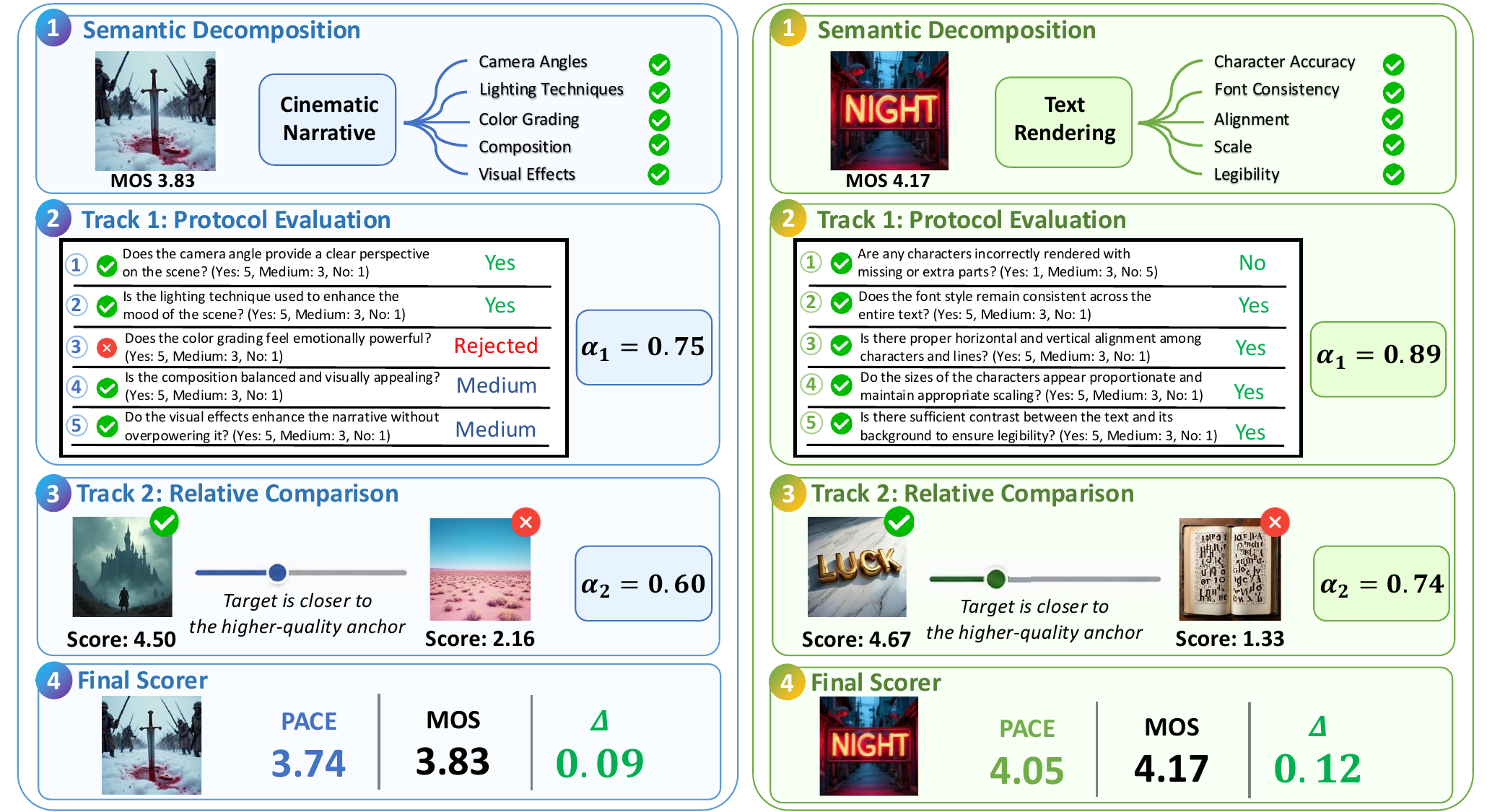}
    \caption{
\textbf{Qualitative case studies of PACE on two open-ended dimensions.}
For each example, PACE first decomposes the target concept into observable sub-dimensions and constructs a Critic-filtered VQA protocol. Track~1 evaluates the target using the resulting protocol, while Track~2 positions it relative to the human-rated high-quality and low-quality anchors. The two continuous perception coefficients, $\alpha_1$ and $\alpha_2$, are computed from logit-weighted probability distributions and fused to obtain the final score. Red crosses indicate rejected protocol components.
}
    \label{fig:example}
\vspace{-0.5cm}
\end{figure}

We present two representative qualitative case studies on the Tier-4 open-ended dimensions in Fig. \ref{fig:example}, \textit{Cinematic Narrative} and \textit{Text Rendering Fidelity}, to illustrate how PACE performs dimension-specific evaluation in practice. For each case, we visualize four stages of the scoring process:
(1) semantic decomposition of the target dimension into observable sub-dimensions,
(2) construction and filtering of the VQA-based evaluation protocol,
(3) dual-track perception using protocol-based absolute evaluation and relative comparison against human-rated calibration anchors, and
(4) fusion into the final quality score.

For \textit{Cinematic Narrative}, the Planning Agent decomposes the concept into camera angles, lighting techniques, color grading, composition, and visual effects. The Critic Agent removes the color-grading rule because it does not provide a sufficiently reliable criterion for the final protocol in this example. The remaining rules produce an absolute perception coefficient of $\alpha_1=0.75$, while comparison with the high- and low-quality anchors gives $\alpha_2=0.60$. Their fusion yields a PACE score of 3.74, close to the human MOS of 3.83.

For \textit{Text Rendering Fidelity}, PACE evaluates character accuracy, font consistency, alignment, scale, and legibility. The resulting protocol produces $\alpha_1=0.89$, while the relative comparison gives $\alpha_2=0.74$. The final prediction is 4.05, compared with a human MOS of 4.17. These examples demonstrate that PACE grounds its predictions in explicit, dimension-specific visual evidence rather than relying on a holistic quality impression.

\subsection{Extensions}
\subsubsection{Limitations.} 
Despite its strong performance, PACE has certain boundaries. First, the cognitive depth of the concept decomposition inherently relies on the reasoning capabilities of the backbone MLLM (e.g., Qwen2.5-VL); if the model lacks the semantic knowledge of an unfamiliar concept, the resulting protocol may be suboptimal. Second, synthesizing visual anchors is bounded by the generation limits of current diffusion models. If the generator fails to produce distinct quality gradations, the Track 2 calibration may become noisy. Lastly, the multi-agent adversarial loop in the Slow-Thinking pipeline incurs higher computational overhead during the initial protocol evolution phase, although this is mitigated by caching the results for Fast-Thinking inference. In addition, the reused protocol is initially grounded on a single image, which provides an efficiency--robustness trade-off but may introduce errors when later images differ substantially from the initial image.

\subsubsection{Broader Impacts.} 
We believe that PACE marks a significant step toward artificial general perception intelligence by decoupling evaluation logic from massive annotated datasets. This democratizes IQA for diverse and fast-evolving AI-generated content (AIGC) applications. This substantially lowers the barrier to establishing fair, unbiased, and customized protocol standards, fostering a healthier and more controllable ecosystem for visual generation technologies.

\end{document}

%% file: math_commands.tex
\usepackage{amsmath,amsfonts,bm}

\def\eqref#1{equation~\ref{#1}}

\def\1{\bm{1}}

\DeclareMathAlphabet{\mathsfit}{\encodingdefault}{\sfdefault}{m}{sl}
\SetMathAlphabet{\mathsfit}{bold}{\encodingdefault}{\sfdefault}{bx}{n}

